\documentclass[letterpaper]{article} 
\usepackage[submission]{aaai23}  
\usepackage{times}  
\usepackage{helvet}  
\usepackage{courier}  
\usepackage[hyphens]{url}  
\usepackage{graphicx} 
\usepackage{natbib}  
\usepackage{caption} 
\usepackage{cite}

\usepackage{subcaption}
\usepackage{amsmath,amssymb,amsfonts}
\usepackage{algorithmicx}
\usepackage[ruled]{algorithm}
\usepackage[noend]{algpseudocode}
\usepackage{graphicx}
\usepackage{textcomp}
\usepackage{xcolor}
\usepackage{paralist}
\usepackage{hyperref}
\usepackage{todonotes,ltablex}
\usepackage{multirow}
\DeclareCaptionStyle{ruled}{labelfont=normalfont,labelsep=colon,strut=off} 
\def\BibTeX{{\rm B\kern-.05em{\sc i\kern-.025em b}\kern-.08em
    T\kern-.1667em\lower.7ex\hbox{E}\kern-.125emX}}

\usepackage{titlesec}

\begin{document}


\title{Towards Adversarial Purification using Denoising AutoEncoders
}

\author{Dvij Kalaria, Aritra Hazra, and Partha Pratim Chakrabarti\\
Dept. of Computer Science and Engineering, Indian Institute of Technology Kharagpur, INDIA}

\maketitle

\begin{abstract}
With the rapid advancement and increased use of deep learning models in image identification, security becomes a major concern to their deployment in safety-critical systems. Since the accuracy and robustness of deep learning models are primarily attributed from the purity of the training samples, therefore the deep learning architectures are often susceptible to adversarial attacks. Adversarial attacks are often obtained by making subtle perturbations to normal images, which are mostly imperceptible to humans, but can seriously confuse the state-of-the-art machine learning models. We propose a framework, named {\bf APuDAE}, leveraging Denoising AutoEncoders (DAEs) to purify these samples by using them in an adaptive way and thus improve the classification accuracy of the target classifier networks that have been attacked. We also show how using DAEs adaptively instead of using them directly, improves classification accuracy further and is more robust to the possibility of designing adaptive attacks to fool them. We demonstrate our results over MNIST, CIFAR-10, ImageNet dataset and show how our framework ({\bf APuDAE}) provides comparable and in most cases better performance to the baseline methods in purifying adversaries. We also design adaptive attack specifically designed to attack our purifying model and demonstrate how our defense is robust to that.

\end{abstract}

\section{Introduction} \label{sec:introduction}

The phenomenal success of deep learning models in image identification and object detection has led to its wider adoption in diverse domains ranging from safety-critical systems, such as automotive and avionics~\cite{rao2018deep} to healthcare like medical imaging, robot-assisted surgery, genomics etc.~\cite{esteva2019guide}, to robotics and image forensics~\cite{yang2020survey}, etc. The performance of these deep learning architectures are often dictated by the volume of correctly labelled data used during its training phases. Recent works~\cite{szegedy2013intriguing}~\cite{goodfellow2014explaining} have shown that small and carefully chosen modifications (often in terms of noise) to the input data of a neural network classifier can cause the model to give incorrect labels. These adversarial perturbations are imperceptible to humans but however are able to convince the neural network in getting completely wrong results that too with very high confidence. Due to this, adversarial attacks may pose a serious threat to deploying deep learning models in real-world safety-critical applications. It is, therefore, imperative to devise efficient methods to thwart such attacks.

Adversarial attacks can be classified into whitebox and blackbox attacks. White-box attacks~\cite{akhtar2018threat} assume access to the neural network weights and architecture, which are used for classification, and thereby specifically targeted to fool the neural network. Hence, they are more accurate than blackbox attacks~\cite{akhtar2018threat} which do not assume access to the model parameters. Many recent works have proposed ways to defend these attacks. Methods for adversarial defense can be divided into 4 categories as -- (i) Modifying the training dataset to train a robust classifier, popularly known as adversarial training, (ii) Block gradient calculation via changing the training procedure, (iii) Detecting adversaries and (iv) purifying adversaries. 
Detecting adversaries only serves half the purpose, what follows logically is to purify the sample or revert it back to its pure form. 
Our main focus is defense mechanisms to purify input data that may have added adversarial perturbation. This can allow the mechanisms to effectively address any attacks. Recent works like MagNet \cite{meng2017magnet}, Defense-GAN \cite{samangouei2018defensegan} consider training a generative model to learn the data distribution closer to the training distribution and map an adversarial example to its corresponding clean example from the data distribution. We follow a similar technique but instead of a generative model like VAE, we use a Denoising AutoEncoder (DAE). We train the DAE specifically to denoise noisy samples instead. Later, during testing, instead of directly passing the input sample which may contain the adversarial noise and treating the later example as a purified example, we use a novel way by adaptively decreasing the reconstruction error to derive the purified sample. This method has two advantages over former method -- (i) Directly passing image through denoising autoencoder often blurs the image as it treats image features as noise as well leading to decreasing target classifier accuracy. Directly modifying the input image to reduce reconstruction error doesn't affect image quality. (ii) The method of purification is non differential so it wouldn't be possible to attack the defense mechanism directly unlike the former method. We observe a significant improvement in results when using DAEs adaptively in comparison to using it directly for MNIST, CIFAR-10 and IMAGENET. 

In summary, the primary contributions made by our work are as follows.
\begin{compactenum}[(a)]
 \item We propose a non-differentiable framework for Adversarial Purification using Denoising AutoEncoder (called {\bf APuDAE}) based on DAE to adaptively purify adversarial attacks which cannot be easily attacked in conjunction in complete white box setting.
 
 \item We test our defense against strong grey-box attacks (attacks where the attacker has complete knowledge of the target classifier model but not the defense model) and attain better than the state-of-the-art baseline methods

 \item We devise possible adaptive attacks specifically designed to attack our method and show how our proposed method is robust to that.

\end{compactenum}
To the best of our knowledge, this is the first work which leverages use of Denoising AutoEncoder architecture adaptively for purifying adversaries to effectively safeguard learned classifier models against adversarial attacks. One of the recent works \cite{Cho2020DAPASD} propose use of DAEs but for defense of image segmentation models instead.

\section{Related Works} \label{sec:literature}




There has been an active research in the direction of adversaries and the ways to avoid them, primarily these methods are statistical as well as machine learning (neural network) based which produce systematic identification and rectification of images into desired target classes. One of the common one used in literature is adversarial training. It particularly aims at retraining or robustifying the target classifier networks against adversarial examples. A comprehensive analysis of adversarial training methods on the ImageNet data set was presented in \cite{he2015deep, Lamb2019InterpolatedAT}. However, one of the major drawback of these methods is that they only make the classifier network robust against specific adversarial attacks but do not generalize well to other attacks. For example, adversarial training done against FGSM does not make the model robust against iterative attacks \cite{carlini2017towards}. Also, they drastically affect the clean accuracy of the target classifier to some extent.
Gradient based methods aim at adding a non differentiable layer at the beginning to make it difficult for the attacker to make gradient based attacks. Basically, they create obfuscated gradients. This may include adding median filter \cite{article_filter}, bit depth reduction \cite{Tran2020SADSD}, random transformation \cite{8954476} etc. However, these kind of attacks can be countered by specifically designing adversaries by either approximating the gradients or using other types of attacks \cite{Athalye2018ObfuscatedGG}.  
Adversarial detection based methods like \cite{hendrycks2016early} \cite{li2017adversarial} \cite{feinman2017detecting} \cite{gao2021maximum} \cite{jha2018detecting} aim at identifying adversaries i.e. detect if the input is an adversary or not. 
Adversarial purification based defenses aim at purifying or adding a new layer in the target classifier architecture to purify the input image to make it closer to the corresponding clean image of the adversary which the target classifier can rightly classify.  Recent works like MagNet \cite{meng2017magnet}, Defense-GAN \cite{samangouei2018defensegan} consider training a generative model to learn the data distribution closer to the training distribution and map an adversarial example to its corresponding clean example from the data distribution as we discussed earlier in Introduction.

\section{Adversarial Attack Models and Methods} \label{sec:background}
For a test example $X$, an attacking method tries to find a perturbation, $\Delta X$ such that $|\Delta X|_k \leq \epsilon_{atk}$ where $\epsilon_{atk}$ is the perturbation threshold and $k$ is the appropriate order, generally selected $\infty$ so that the newly formed perturbed image, $X_{adv} = X + \Delta X$. Here, each pixel in the image is represented by the ${\tt \langle R,G,B \rangle}$ tuple, where ${\tt R,G,B} \in [0, 1]$. In this paper, we consider only white-box attacks, i.e. the attack methods which have access to the weights of the target classifier model. 

\subsection{Random Perturbation (RANDOM)}
Random perturbations are simply unbiased random values added to each pixel ranging in between $-\epsilon_{atk}$ to $\epsilon_{atk}$. Formally, the randomly perturbed image is given by,
\begin{equation}
X_{rand} = X + \mathcal{U}(-\epsilon_{atk},\epsilon_{atk})
\end{equation}
where, $\mathcal{U}(a,b)$ denote a continuous uniform distribution in the range $[a,b]$.

\subsection{Fast Gradient Sign Method (FGSM)}
Earlier work by~\cite{goodfellow2014explaining} introduced the generation of malicious biased perturbations at each pixel of the input image in the direction of the loss gradient $\Delta_X L(X,y)$, where $L(X,y)$ is the loss function with which the target classifier model was trained. Formally, the adversarial examples with $\epsilon_{atk}$ are computed as :-
\begin{equation}
X_{adv} = X + \epsilon_{atk} . sign(\Delta_X L(X,y))
\end{equation}

\subsection{Projected Gradient Descent (PGD) or BIM}
Earlier works by~\cite{Kurakin2017AdversarialML} propose a simple variant of the FGSM method by applying it multiple times with a rather smaller step size than $\epsilon_{atk}$. However, as we need the overall perturbation after all the iterations to be within $\epsilon_{atk}$-ball of $X$, we clip the modified $X$ at each step within the $\epsilon_{atk}$ ball with $l_\infty$ norm. 
\begin{subequations}
\begin{flalign}
& X_{adv,0} = X,\\
& X_{adv,n+1} = {\tt Clip}_X^{\epsilon_{atk}}\Big{\{}X_{adv,n} + \alpha.sign(\Delta_X L(X_{adv,n},y))\Big{\}}
\end{flalign}
\end{subequations}
Given $\alpha$, we take the no. of iterations, $n$ to be $\lfloor \frac{2 \epsilon_{atk}}{\alpha}+2 \rfloor$. This attacking method has also been named as Basic Iterative Method (BIM) in some works.

\subsection{Carlini-Wagner (CW) Method}
\cite{carlini2017towards} proposed a more sophisticated way of generating adversarial examples by solving an optimization objective as shown in Equation~\ref{carlini_eq}. Value of $c$ is chosen by an efficient binary search. We use the same parameters as set in \cite{li2020deeprobust} to make the attack.
\begin{equation} \label{carlini_eq}
X_{adv} = {\tt Clip}_X^{\epsilon_{atk}}\Big{\{}\min\limits_{\epsilon} \left\Vert\epsilon\right\Vert_2 + c . f(x+\epsilon)\Big{\}}
\end{equation}



\section{Our Proposed {\bf APuDAE} Framework} \label{sec:method}
In this section, we present our proposed framework called {\em Adversarial Purification using Denoising AutoEncoders} ({\bf APuDAE}) which shows how Denoising AutoEncoders (DAE), trained over a dataset of clean images, are capable of purifying adversaries.

\subsection{AutoEncoders (AE)}
AutoEncoder is a type of artificial neural network used to learn efficient representations for unlabelled data using unsupervised learning. It has two components known as Encoder and Decoder. The encoder maps the high dimensional image input to a low dimension latent space with latent dimension $d$ and decoder maps the latent representation back to the reconstructed input with same size as the image. The encoder and decoder layers are trained with the objectives to get the reconstructed image as close to the input image as possible thus forcing to preserve most of the features of the input image in the latent vector to learn a compact representation of the image. 

\begin{figure}[h] 
    \centering
    \includegraphics[width=0.5\textwidth]{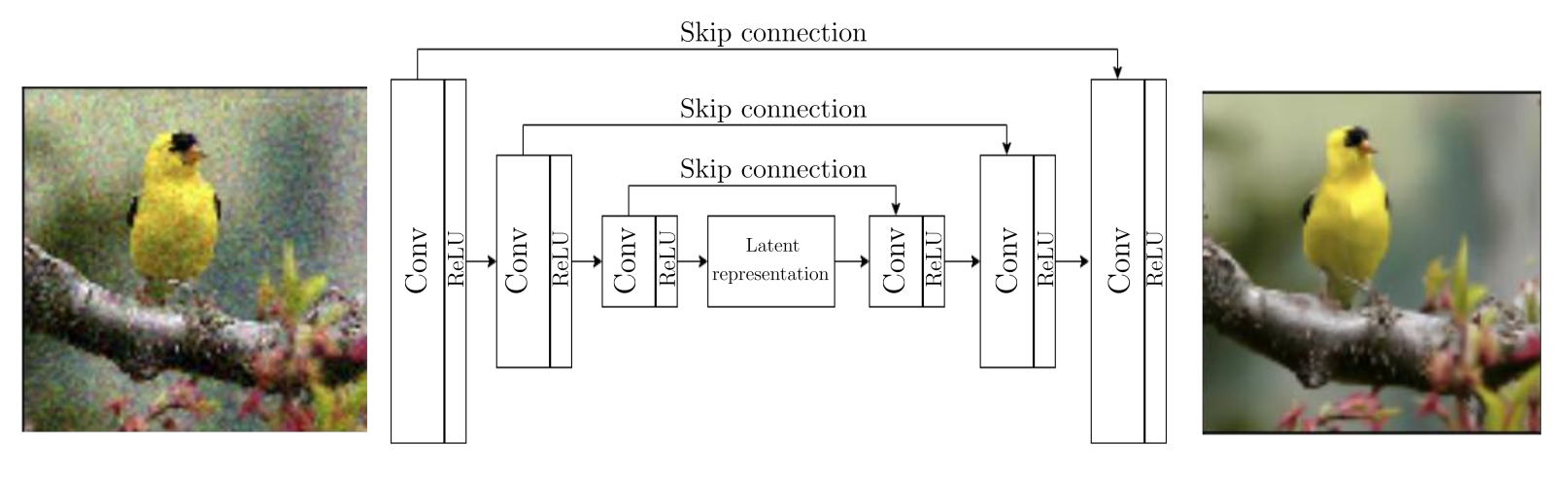}
    \caption{DAE Model Architecture}
    \label{fig:cvae_diag}
\end{figure}

Denoising AutoEncoder (DAE) is a variation of autoencoder in which the input is passed as a noisy image and the output has to be a clean image. The encoder learns to map noisy image to a latent representation corresponding to a clean image and decoder maps it back to the clean image. Training is carried out in unsupervised manner with the input image obtained by adding random gaussian noise with varying magnitudes and the output image is the clean image itself. The gaussian noise added is obtained with varying values of variance, $\sigma$. $\sigma$ is chosen from a uniform distribution $[0,\sigma_{max}]$ so that the model is not biased to remove noise of specific magnitude. Formally, the loss functions is defined as follows where the $X_{noise}$ is the added noise, $F_{dae}(.)$ is the trained DAE which takes and outputs an image, mse is the mean square error loss.
\begin{equation} \label{eqn:inp_eq}
\begin{split}
    &L = mse(F_{dae}(X_{inp}), X_{im}) \\
    &\text{where, } X_{inp} = X_{im} + \sigma X_{noise},\\
    &\text{and } \sigma \sim U[0,\sigma_{max}],\ \ x_{noise} \sim \mathcal{N}(0,1)\ \ (\forall x_{noise} \in X_{noise})
\end{split}
\end{equation}

\subsection{Using Skip Connections}
Deep neural networks suffer from the degradation problem i.e. some information is lost after each convolution or deconvolution step. Residual networks involving skip connections are a known solution to this problem. These additional connections directly send the feature maps from an earlier layer to the later corresponding layer of the decoder. This helps the decoder form more clearly defined decompositions of the input image. Mathematically, we define skip connection as follows, where $X_{inp}$ is the input image, $X_{enc,i}$ is the $i^{th}$ feature map for the encoder and $X_{dec,i}$ is the $i^{th}$ feature map for the decoder layer, $F_{enc,i}$ is the $i^{th}$ encoder layer comprising of a convolution operation followed by a dropout and an activation layer, and $F_{dec,i}$ is the $i^{th}$ decoder layer comprising of a deconvolution operation followed by a dropout and an activation layer except for the last layer, and $N$ is the no. of layers :-

\begin{equation} \label{eq:skip_conn}
\begin{split}
    &X_{enc,i+1} = F_{enc,i}(X_{enc,i})\\ 
    &X_{dec,i-1} = F_{dec,i}(X_{dec,i}) + X_{enc,i-1} \text{for } i \in {0,1...N-1} \\ 
\end{split}
\end{equation}

The layers sizes are tabulated in Table~\ref{tab:cvae_arch_sizes}. The network architecture used for ImageNet is relatively more complex and is the same as proposed in \cite{Laine2019HighQualitySD}, hence the readers are referred to it for more details. 

\begin{table}[h]
{\sf \footnotesize
\begin{center}
\begin{tabular}{|c||c|c|}
    \hline
  {\bf Attribute}  & {\bf Layer}     & {\bf Size}    \\
  \hline
  \hline
          & Conv2d+Relu+BatchNorm2d      & (3,2,1,64)\\ 
          \cline{2-3}
{\bf Encoder}   & Conv2d+Relu+BatchNorm2d      & (3,1,1,64)\\ 
          & $\times (N-1)$ & \\
  \hline
          & DeConv2d+Relu+BatchNorm2d      & (3,1,1,64)\\ 
          & $\times (N-1)$ & \\ 
          \cline{2-3}
{\bf Decoder}   & DeConv2d+Relu+BatchNorm2d      & (3,2,1,c)\\ 
          \cline{2-3}
          & Sigmoid     &   \\
\hline
\end{tabular}
\end{center}
}
\caption{DAE Architecture Layer Sizes. $c$ = Number of Channels in the Input Image ($c=3, N=15$ for CIIFAR-10 and $c=1,N=5$ for MNIST). Size format : (Kernel size, stride, padding, no. of output channels).}
\label{tab:cvae_arch_sizes}
\end{table}



\subsection{Determining Reconstruction Errors}
\begin{figure}[h] 
    \begin{subfigure}{.115\textwidth}
        \centering
        \includegraphics[width=\textwidth]{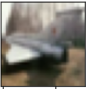}
        \caption{airplane}
    \end{subfigure}
    \begin{subfigure}{.115\textwidth}
        \centering
        \includegraphics[width=\textwidth]{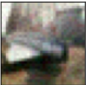}
        \caption{truck}
    \end{subfigure}
    \begin{subfigure}{.115\textwidth}
        \centering
        \includegraphics[width=\textwidth]{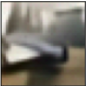}
        \caption{ship}
    \end{subfigure}
    \begin{subfigure}{.115\textwidth}
        \centering
        \includegraphics[width=\textwidth]{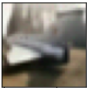}
        \caption{airplane}
    \end{subfigure}
    \caption{Difference in purifications from directly passing through DAE and using adaptive method. Left to right : Input Image, Adversarial image, Directly purified image, Purified image by adaptive method. The sub-labels are the inferred classes from the target classifier network}
    \label{fig:demon_eg}
\end{figure}

Let $X$ be the input image and $X_{rcn}$ is the reconstructed image obtained from the trained DAE. We define the reconstruction error or the reconstruction distance as in Equation~\ref{eq:recon}.
\begin{equation} \label{eq:recon}
    {\tt Recon}(X) = (X - X_{rcn})^2
\end{equation}
Two pertinent points to note here are:
\begin{compactitem}
    \item For clean test examples, the reconstruction error is bound to be less since the DAE is trained with unknown noise magnitude, hence DAE should ideally leave the input image unaffected. However, some image features may be perceived as noise leading to slight reconstruction distance for the clean images
    
    \item For the adversarial examples, as they contain the adversarial noise, passing the adversarial image leads to removal of some part of the adversarial noise, hence leading to high reconstruction distance as the output image does not ideally contain the adversarial noise. 
\end{compactitem}
We use this peculiar difference to our advantage by modifying the input image itself so that the reconstruction error reduces and leads to a clean image. We do this instead of directly taking the output image from DAE as the purified image i.e. $X_{pur} := X_{rcn}$. Modifying the input image to reduce reconstruction has the following advantages over the later method :- 1) On passing the input image directly through the DAE, some of the image features are lost, leading to reduced target classifier accuracy. Hence using the former method of modifying the input image itself leads to better image quality. 
2) The purification function is non differentiable in the first case while the later method is differentiable as the DAE is differentiable, hence it becomes easy for the attacker to create an adaptive attack by attacking both the target classifier and DAE. 
As an example, let the clean image be an image of a airplane and its slightly perturbed image fools the classifier network to believe it is a truck. Hence, the input to the DAE will be the slightly perturbed airplane image with the predicted class truck. 
Now, on passing the perturbed image directly through DAE, adversarial perturbations are removed but some image features are also perceived as noise and also removed. This leads to target classifier misclassifying the modified image as a ship (see Figure~\ref{fig:demon_eg}). 

\alglanguage{pseudocode}
\begin{algorithm}
\small
\caption{Adversarial Purification Algorithm}
\label{algo:adv_detect}
\begin{algorithmic}[1]
\Function{purify ($dae, x, recon_dists$)}{}
    \For{$n$ times}
        \State $x$ $\gets$ $x$ - $\alpha$ $\frac{\partial dae(x)}{\partial x}$ 
    \EndFor
    \State {\bf return} x
\EndFunction
\Function{Classify\_Purify\_Adversaries ($X_{train}, Y_{train}, X, classifier$)}{}
    \State dae $\gets$ ${\tt Train}(X_{train})$
    \State recon\_dists $\gets$ ${\tt dae}(X_{train})$
    \State Y $\gets$ $\phi$
    \For{$x$ in $X$}
        \State $X_{pur}$ $\gets$ ${\tt PURIFY}($dae,$x$,recon\_dists$)$
        \State $y_{pred}$ $\gets$ ${\tt classifier}(X_{pur})$
        \State Y.${\tt insert}(y_{pred})$
    \EndFor
    \State {\bf return} Y 
\EndFunction
\Statex
\end{algorithmic}
\end{algorithm}

\section{Experimental Results} \label{sec:experiment}

\subsection{MNIST Dataset}

We present the comparison in results for MNIST dataset \cite{lecun2010mnist} with the 2 methods discussed earlier MagNet \cite{meng2017magnet} and DefenseGAN \cite{samangouei2018defensegan}(see Table \ref{tab:comp_res}). Results for adversarial training are generated in the same way as described in MagNet. All results are with $\epsilon=0.1$ as used commonly in most works \cite{samangouei2018defensegan, Madry2018TowardsDL}. We use no. of iterations for purification, $n=15$ and $\alpha=0.01$.

\setlength\tabcolsep{1.5pt}
\begin{table}[h]
{\sf \scriptsize
\begin{center}
\begin{tabular}{|c|c|c|c|c|c|c|}
    \hline
  {\bf Attack}  & {\bf No}  & {\bf Adversarial} & {\bf Defense-} & {\bf MagNet} & {\bf APuDAE} & {\bf APuDAE}  \\
  {($\epsilon$=$0.1$)}  & {\bf Defense}  & {\bf Training} & {GAN} & {} & {(Direct)} & {($n$=$15$,$\alpha$=$0.01$)}   \\
  \hline \hline
    Clean  &  0.974 & 0.822 & 0.969 & 0.95 & 0.973 & {\color{red} 0.973}\\ 
  \hline
    FGSM  &  0.269 & 0.651 & 0.949 & 0.69 & 0.934 & {\color{red} 0.964}\\ 
  \hline
    R-FGSM  &  0.269 & 0.651 & 0.945 & 0.65 & 0.921 & {\color{red} 0.962}\\ 
  \hline
    PGD  &  0.1294 & 0.354 & 0.939 & 0.49 & 0.912 & {\color{red} 0.951} \\ 
  \hline
    CW  & 0. & 0.28 & 0.936 & 0.45 & 0.907 & {\color{red} 0.953}\\ 
\hline
\end{tabular}
\end{center}
}
\caption{Comparison of Results using {\bf MNIST} Dataset. Values reported are classifier success rate (in fraction out of 1). {\bf APuDAE (Direct)}: When the image directly obtained from DAE is used. {\bf APuDAE}: When back propagation through reconstruction error is used for purification}
\label{tab:comp_res}
\end{table}

\subsection{CIFAR-10 Dataset}
MNIST dataset is a simple dataset with easily identifiable number shapes which can be robustly learned by a neural network. It is therefore very easy to correct a perturbed image with a simple back-propagation method with fixed no. of iterations. For CIFAR-10 dataset, due to the inherent complexity in classifying objects, it is very easy to attack and thus simple defense with fixed no. of iterations is not enough to give a reasonably good purification result, hence we explore the following variations in algorithm.

\subsubsection{Fixed Number of Iterations.}

First, we evaluate the results with fixed no. of iterations. The results are listed on Table \ref{tab:comp_cifar}. As can be observed with more no. of iterations clean accuracy gets reduced and adversarial accuracy improves. The drawback with such method is that it is not essentially using any discrimination for the update rule for adversarial and non adversarial examples. As a result of this, reconstruction errors of both adversarial and clean images (see Figure \ref{fig:recons_errors_A}) are reduced leading to disappearance of important features. Formally, the purified example, $X_{pur}$ can be expressed as follows where $X_{adv}$ is the adversarial image, $n$ is the fixed no. of iterations, $\alpha$ is the constant learning rate and $purifier$ is the Purifier autoencoder function. Comparison of results on varying no. of iterations for purifying adversaries is presented in technical appendix attached to this submission. 
\begin{equation}
\begin{split}
    &X_{pur,0} = X_{adv} \\
    &X_{pur,i+1} = X_{pur,i} - \alpha \frac{\partial L(X_{pur,i},purifier(X_{pur,i}))}{\partial X_{pur,i}} \\
    &\text{for } i \in \{1,2...,n\} \text{ and } L(X,Y) = (X-Y)^2
\end{split} \label{Eqn:fix_iter_basic}
\end{equation}
\subsubsection{Fixed Number of Iterations with ADAM for Update.}

To counter the drawback from using fixed no iterations, we use ADAM to achieve the minima quickly. In this case, the update amount varies for adversarial and clean image as clean image has less reconstruction error (see Figure \ref{fig:recons_errors_B}). Formally, the update is defined as follows where $X_{adv}$ is the adversarial image, $X_{pur}$ is the purified image, $n$ is the no. of iterations, $\alpha$ and $\beta$ are parameters for ADAM optimizer.
\begin{equation}
\begin{split}
    &X_{pur,0} = X_{adv} \\
    &X_{pur,i+1} = X_{pur,i} - \alpha w_i\\
    &\text{where, } w_i = \beta w_{i-1} + (1-\beta) \frac{\partial L(X_{pur,i},purifier(X_{pur,i}))}{\partial X_{pur,i}} \\
    &\text{for } i \in \{1,2...,n\} \text{ and } L(X,Y) = (X-Y)^2
\end{split} \label{Eqn:fix_iter_adam}
\end{equation}

\subsubsection{Variable Learning Rate based on Current Reconstruction Error.}

Usually for adversarial images, the initial reconstruction error is high as compared to clean image. Based on the following, an advantage can be taken to purify mostly only the adversarial images and not the clean images. Learning rate can be varied based on the reconstruction error of the input sample. First, the mean value, $\mu$ and variance, $\sigma$ of the distribution of reconstruction errors for the clean train samples are determined. Then, the probability of falling within the equivalent gaussian distribution of reconstruction distances of clean validation samples is determined. The learning rate is varied accordingly linear to this probability. Formally, update rule can be defined as follows where $X_{adv}$ is the adversarial image, $X_{pur}$ is the purified image, $n$ is the no. of iterations, $\alpha$ and $\beta$ are parameters for ADAM optimizer.
\begin{equation}
\begin{split} \label{Eqn:fix_iter_adam_variable_lr}
    &X_{pur,0} = X_{adv} \\
    &X_{pur,i+1} = X_{pur,i} - \alpha_i w_i \\
    &\text{where, } \alpha_i = \alpha (1 - \exp^{-(\frac{X_{pur,i}-\mu}{\sigma})^2}), \\
    &\text{and } w_i = \beta w_{i-1} + (1-\beta) \frac{\partial L(X_{pur,i},purifier(X_{pur,i}))}{\partial X_{pur,i}} \\
    &\text{for } i \in \{1,2...,n\} \text{ and } L(X,Y) = (X-Y)^2
\end{split} 
\end{equation}

\subsubsection{Set Target Distribution for Reconstruction Error.}

Major drawback with earlier variations is that even though they discriminate in purifying already clean and adversarial images, they still try to purify clean images by bringing down the reconstruction error. One drawback to this is on doing so, the important features of the image are lost as the purification model sees them as perturbations to the images giving rise to the reconstruction error. Back-propagating through them smoothens them leading to less reconstruction error but confuses the classifier leading to wrong predictions. To avoid this, we keep the objective for the update rule to increase the probability mass function value with respect to the target distribution. Formally, it is defined as;
\begin{equation}
\begin{split}
    &X_{pur,0} = X_{adv} \\
    &X_{pur,i+1} = X_{pur,i} - \alpha w_i\\
    &\text{where, } w_i = \beta w_{i-1} + (1-\beta) \frac{\partial L(X_{pur,i},purifier(X_{pur,i}))}{\partial X_{pur,i}} \\
    &\text{for } i \in \{1,2...,n\} \text{ and } \\
    &L(X,Y) = \frac{|dist(X,Y)-\mu|}{\sigma},\ dist(X,Y) = (X-Y)^2
\end{split} \label{Eqn:fix_iter_target}
\end{equation}

\subsubsection{Set Target Distribution for Reconstruction Error with Modified Update Rule.}

\begin{figure}[h!] 
    \begin{center}
    \begin{subfigure}{.23\textwidth}
        \centering
        \includegraphics[width=\textwidth]{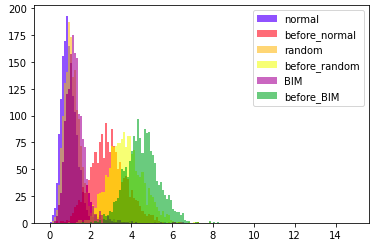}
        \caption{Fixed no. of iterations}
        \label{fig:recons_errors_A}
    \end{subfigure}
    \begin{subfigure}{.23\textwidth}
        \centering
        \includegraphics[width=\textwidth]{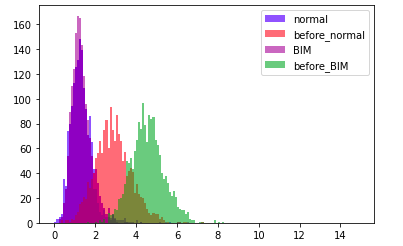}
        \caption{Fixed no. of iterations with using ADAM optimizer}
        \label{fig:recons_errors_B}
    \end{subfigure}
    \begin{subfigure}{.23\textwidth}
        \centering
        \includegraphics[width=\textwidth]{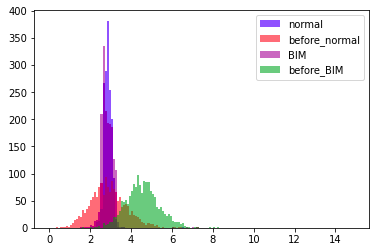}
        \caption{Set target distribution for reconstruction error}
        \label{fig:recons_errors_C}
    \end{subfigure}
    \begin{subfigure}{.23\textwidth}
        \centering
        \includegraphics[width=\textwidth]{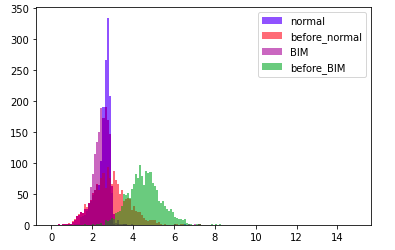}
        \caption{Set target distribution for reconstruction error with modified update rule}
        \label{fig:recons_errors_D}
    \end{subfigure}
    \caption{Reconstruction errors for different variations}
    \label{fig:recon_errors_mnist}
    \end{center}
\end{figure}

Drawback of the above method is that it tries to increase the reconstruction error (see Figure \ref{fig:recons_errors_C}) of the samples with less reconstruction error belonging to clean image set. Due to this, classification model gives less accuracy. To avoid this, we change the update rule by just modifying the loss function, $L(X,Y) = \frac{max(dist(X,Y)-\mu,0)}{\sigma}$ which ultimately leads to no change for clean images with reconstruction error less than $\mu$. 

\subsubsection{Adding Random Noise at Each Update Step.}\label{best_results}

Just adding random noise has been observed to improve the classification accuracy for adversarial examples as it changes the overall perturbation to near random breaking the targeted perturbation created by the adversarial attack to the corresponding clean sample. This technique can be used in conjunction to our method where random perturbation is added at each update step. This leads to slightly better results as observed in Table \ref{tab:comp_cifar}. Formally the update rule is defined as follows where $\gamma$ is the amount of noise to be added at each update step.
\begin{equation}
\begin{split}
    &X_{pur,0} = X_{adv} \\
    &X_{pur,i+1} = X_{pur,i} - \alpha w_i\\
    &\text{where, } w_i = \beta w_{i-1} + \\
    & \qquad \qquad \qquad (1-\beta) \frac{\partial L(X_{pur,i},purifier(X_{noise,pur,i}))}{\partial X_{noise,pur,i}} \\
    &\text{for } i \in \{1,2...,n\} \text{ and } \\
    &X_{noise,pur,i} = X_{pur,i} + \gamma r_X, r_X ~ \mathcal{N}(0,I_X), \\
    &L(X,Y) = \frac{max(dist(X,Y)-\mu,0)}{\sigma}, \\
    &dist(X,Y) = (X-Y)^2
\end{split} \label{Eqn:fix_iter_target_random}
\end{equation}

\subsubsection{Add random transformation before}

Adding random rotate and resize transformations to the input image has also been observed to improve the classification accuracy for adversarial examples \cite{8954476}. This technique can be used in conjunction to our method where random transformation can be added at the beginning. This however lead to slightly worse results as observed in Table \ref{tab:comp_cifar} as the distortion in the low resolution image caused by transformation over-weighed the advantage from apparent destruction of adversary pattern at each iteration. Hence, the results obtained from adding random noise at each step is observed to be the best in the case of Cifar-10. Formally the update rule is defined as follows where $t$ is the transformation function taking the resize factor $f$ and rotation $\theta$ as input.
\begin{equation}
\begin{split}
    &X_{pur,0} = t(X_{adv},f,\theta) \\
    &X_{pur,i+1} = X_{pur,i} - \alpha w_i\\
    &\text{where, } w_i = \beta w_{i-1} + \\
    &\qquad \qquad \qquad (1-\beta) \frac{\partial L(X_{pur,i},purifier(X_{trans,pur,i}))}{\partial X_{trans,pur,i}} \\
    &\text{for } i \in \{1,2...,n\} \text{ and } \\
    &L(X,Y) = \frac{max(dist(X,Y)-\mu,0)}{\sigma}, \\
    &dist(X,Y) = (X-Y)^2
\end{split} \label{Eqn:fix_iter_target_trans}
\end{equation}

\setlength\tabcolsep{1pt}
\begin{table}[h!]
{\small
\begin{center}
\begin{tabular}{|c||c|c|c|c|c|c|}
\hline
{\bf Attack}               & {\bf Clean}                       & {\bf Random}                     & {\bf FGSM}                        & {\bf R-FGSM}                      & {\bf BIM}        & {\bf CW}                                      \\ \hline \hline
{\bf No Defense}           & 0.954                       & 0.940                       & 0.533                       & 0.528                       & 0.002   &   0                          \\ \hline
{\bf Adversarial} & 0.871                       & 0.865                          & 0.650                       & 0.640                       & 0.483       & 0.412            \\ 
{\bf Training} &                        &                           &                        &                       &                &                             \\ \hline
{\bf APuDAE}               & 0.790                       & 0.793                      & 0.661                       & 0.655                       & 0.367       &   0.351 \\ 
{\bf (Direct)}           &                        &                       &                         &                         &        &  \\ \hline
{\bf APuDAE}           & 0.879                       & 0.889                      & 0.73                        & 0.73                        & 0.632       & 0.621 \\ 
{\bf (A, $n$=12)}           &                        &                       &                         &                         &        &  \\ \hline
{\bf APuDAE}           & 0.853                       & 0.875                      & 0.76                        & 0.747                       & 0.694   & 0.683                                         \\ 
{\bf (A, $n$=15)}           &                        &                       &                         &                         &        &  \\ \hline
{\bf APuDAE}           & 0.843                       & 0.867                      & 0.774                       & 0.764                       & 0.702       & 0.695                                  \\ 
{\bf (A, $n$=18)}           &                        &                       &                         &                         &        &  \\ \hline
{\bf APuDAE(B)}                    & 0.872                       & 0.883                      & 0.778                       & 0.772                       & 0.706       & 0.698                              \\ \hline
{\bf APuDAE(C)}                    & 0.892                       & 0.874                      & 0.769                       & 0.762                       & 0.692       & 0.689             \\ \hline
{\bf APuDAE(D)}                    & 0.904 & 0.889 & 0.783                       & 0.767                       & 0.684                                     & 0.679       \\ \hline
{\bf APuDAE(E)}                    & 0.907                       & {\color{red} 0.893}                      & {\color{red} 0.834} & {\color{red} 0.817} & 0.702   & 0.689  \\ \hline
{\bf APuDAE(F)}                    & {\color{red} 0.908}                       & 0.893                      & 0.818                       & 0.811                       & {\color{red} 0.729}               & {\color{red} 0.708}                             \\ \hline
{\bf APuDAE(G)}                    & 0.858                       & 0.846                      & 0.828                       & 0.814                       & 0.711                   & 0.700                         \\ \hline
\end{tabular}
\end{center}
}
\caption{Comparison of Results for Different Variations on {\bf CIFAR-10} dDataset. {\bf Direct}: Direct usage of DAE, {\bf A}: Fixed no. of iterations, {\bf B}: Fixed no. of iterations with using ADAM optimizer, {\bf C}: Variable learning rate based on the current reconstruction error, {\bf D}: Set target distribution for reconstruction error, {\bf E}: Set target distribution for reconstruction error with modified update rule, {\bf F}: Add random noise at each update step, {\bf G}: Add random transformation at the beginning}
\label{tab:comp_cifar}
\end{table}

\vspace{-0.35cm}
\subsection{ImageNet Dataset}

ImageNet dataset is the high resolution dataset with cropped RGB images of size 256X256. For our experiments we use pretrained weights on ImageNet train corpus available from \cite{5206848}. For test set we use the 1000 set of images available for ILSVRC-12 challenge \cite{Russakovsky2015ImageNetLS} which is commonly used by many works \cite{xu2017feature} for evaluating adversarial attack and defense methods. This test set consists of 1000 images each belonging to a different class. The pre-trained weights of ResNet-18 classifier are also provided by ILSVRC-12 challenge \cite{Russakovsky2015ImageNetLS}. The 1000 images are chosen such that the classifier gives correct class for all of them. 

The classification accuracy for ImageNet dataset comprising of 1000 classes is defined by the top-1 accuracy which means a prediction is correct if any of the top 1\% of total classes or 10 class predictions in this case are correct. Hence, for the iterative attack (which is used as baseline by all recent works) $J(.)$ is the cross entropy loss. Effect of varying the no. of iterations of update for purifying adversarial examples can be seen in Table \ref{tab:comp_imagenet}. As can be observed, increasing no. of iterations lead to better classification accuracy for purified adversaries but leads to drop in clean accuracy. Results for adversaries created with different no. of iterations are also reported. The value of $\epsilon=\frac{8}{255}$ and update step, $\alpha=\frac{1}{255}$ are chosen as standard values similar to ones used in previous literature \cite{xu2017feature}. We further explore different variations to the base version with fixed no. of iterations and constant update rate $\alpha$.

\subsubsection{Random Noise at Each Step.}

Similar to Cifar-10 dataset, adding random noise at each update step leads to increase in classification accuracy. As evident from Table \ref{tab:comp_imagenet}, there is a slight surge in classification accuracy of purified adversaries with this addition. 


\subsubsection{Random Transformation at Each Step.}

Similar to Cifar-10 dataset, we add random rotate and resize transformation before the purification. This too results in a subsequent increase in classification accuracy of purified adversaries as observed in Table \ref{tab:comp_imagenet}. The reason is that the transformation leads to change in the locality of the features where targeted attack was made. The classification network however has been trained to act robustly if the image is resized or rotated, hence the classified class for clean images doesn't change, while for attacked images due to relocation of targeted perturbations, the resultant perturbation tends towards non-targeted random noise leading to increased accuracy. Please refer to attached technical appendix for comparison with different values of transformation parameters.

\setlength\tabcolsep{1pt}
\begin{table}[h]
{\small
\begin{center}
\begin{tabular}{|c||c|c|c|c|}
\hline
{\bf Attack}   & {\bf Clean}            & {\bf Random}              & {\bf BIM} & {\bf BIM} \\  
 &   &  & {\bf \big{(}$\epsilon=\frac{5}{255}$ \big{)}} & {\bf \big{(}$\epsilon=\frac{25}{255}$ \big{)}} \\ \cline{1-5} 
{\bf No defense}  & 1.0                       & 0.969                     & 0.361                     & 0.002                                            \\
\hline \hline
{\bf Adv. Training} & 0.912            & 0.901                     & 0.641                     & 0.454                                            \\
\hline
{\bf APuDAE (Direct)}      & {\color{red} 0.946}               & {\color{red} 0.941}               & 0.909                     & 0.899                                        \\
\hline
{\bf APuDAE (A,$n=12$)}  & 0.941                     & 0.936                     & 0.919                    & 0.911                                           \\ \cline{1-5} 
{\bf APuDAE (A,$n=15$)}  & 0.939                     & 0.933                     & 0.917                     & 0.914                                            \\
\hline
{\bf APuDAE (A,$n=18$)}  & 0.934                     & 0.932                     & 0.917                     & 0.913                                            \\
\hline
{\bf APuDAE (B,$n=15$)}  & 0.938                     & 0.930                     & 0.923                    & 0.920                                            \\
\hline
{\bf APuDAE (C,$n=15$)}  & 0.926                     & 0.924                     & {\color{red} 0.928}               & {\color{red} 0.929}                                       \\
\hline
\end{tabular}
\end{center}
}
\caption{Comparison of Results on Imagenet Dataset. {\bf Direct}: Direct usage of DAE, {\bf A}: No variation, {\bf B}: With random noise, {\bf C}: With random transformations. $\epsilon$ is the magnitude of adversarial perturbation and $n$ is the no. of iterations of APuDAE for purification.}  
\label{tab:comp_imagenet}
\end{table}

\vspace{-0.2cm}
\section{Possible Counter Attacks}

Based on the attack method, adaptive attacks can be developed to counter the attack. We study the 2 types of attacks in detail. For a detailed review on how to systematically design an adaptive counter attack based on the category of defense, readers are referred to \cite{Laine2019HighQualitySD}.

\subsection{Counter Attack A}

The first adaptive attack possible is designed by approximating the transformation function i.e. the function of the output image obtained by modifying the input image through the update rule, by the differentiable function obtained by autoencoder end-to-end network. Intuitively this can be thought of as the output obtained by backpropagating through the reconstruction error loss between the input and purified output after passing through autoencoder purifier network. We observe (see Table \ref{tab:comp_counter}) that on applying this counter attack, the accuracy for direct purifier output method drops drastically while ours method gives better robust results against this attack. Mathematically, we can express the attacked image, $X_{attacked}$ as follows where $X_{cln}$ is the original clean image, $n$ is the no. of iterations, $classifier(.)$ is the target classifier network, $J(.)$ is the cross entropy loss, $purifier(.)$ is the purifier autoencoder, $\alpha$ is the update rate and $\pi_{X,\epsilon}(.)$ function restricts value within $[X-\epsilon,X+\epsilon]$.
\begin{equation}
\begin{split}
    &X_{attacked,0} = X_{cln} \\
    &X_{attacked,i+1} = \pi_{X_{cln},\epsilon}[X_{attacked,i} + \\
    &\qquad \qquad \qquad \alpha. sign(\frac{\partial J(classifier(X_{attacked,i}),y)}{\partial X_{attacked,i}})] \\
    &\text{ for } i \in \{1,2...,n\}
\end{split} \label{Eqn:counter_attack_A}
\end{equation}

\subsection{Counter Attack B}

The second adaptive attack possible here is by modifying the loss function of the BIM attack by including new weighted term getting less reconstruction error from the purifier This way we attack both the classifier as well as purifier. The purifier method relies on reconstruction error as a measure to update the input to get less reconstruction error similar to clean images but if the attack is made with this consideration to fool the purifier method by giving similar reconstruction error to clean image while also fooling the classifier, the attack is successful as it bypasses the purifier method. For this attack, the attacked image seems to be attacked more at the edges as modifying those do not change the reconstruction error for purifier much. As observed from Table \ref{tab:comp_counter}, the adaptive counter attack is successful to an extent but still performs considerably better than adversarial training. Mathematically, we can express the attacked image, $X_{attacked}$ as follows where $X_{cln}$ is the original clean image, $n$ is the no. of iterations, $classifier(.)$ is the target classifier network, $J(.)$ is the cross entropy loss, $purifier(.)$ is the purifier autoencoder, $\alpha$ is the update rate, $\beta$ is the weighing factor for the reconstruction error in the combined loss function, and $\pi_{X,\epsilon}(.)$ function restricts value within $[X-\epsilon,X+\epsilon]$. The value of $\beta$ is chosen to get worst possible attack for each dataset from this scheme. 
\begin{equation}
\begin{split}
    &X_{attacked,0} = X_{cln} \\
    &X_{attacked,i+1} = \pi_{X_{cln},\epsilon}[X_{attacked,i} + \alpha sign(\\
    &\ \ \ \ \ \ \ \ \ \ \ \ \ \ \ \ \ \ \ \ \ \ \ \ \ \ \  \frac{\partial }{\partial X_{attacked,i}}(J(X_{attacked,i},y) + \\
    &\ \ \ \ \ \ \ \ \ \ \ \ \ \ \ \ \ \ \ \ \ \ \ \ \ \ \ \beta (X_{attacked,i} - purifier(X_{attacked,i}))^2)] \\
    &\text{ for } i \in \{1,2...,n\}
\end{split} \label{Eqn:counter_attack_B}
\end{equation}

\setlength\tabcolsep{2.5pt}
\begin{table}[h]
{\small
\begin{center}
\begin{tabular}{|c||c|c|c|c|}
\hline
\textbf{Dataset} & \textbf{Counter}  & \textbf{Adversarial} & \textbf{APuDAE} & \textbf{APuDAE}  \\
\textbf{} & \textbf{Attack}  & \textbf{Training} & {\bf (Direct)} & {\bf (Best)} \\ \hline \hline
\textbf{MNIST} & A   & 0.354                         & 0.799           & {\color{red} 0.891} \\ \cline{2-5} 
 & B & 0.354                         & 0.815           & {\color{red} 0.857} \\ \hline
{\textbf{Cifar-10}} & A & 0.483                         & 0.199           & {\color{red} 0.680} \\ \cline{2-5} 
 & B & 0.483                         & 0.275           & {\color{red} 0.507} \\ \hline
\textbf{ImageNet}                       & A & 0.254                         & 0.134           & {\color{red} 0.536} \\ \cline{2-5} 
                                                         & B & 0.254                         & 0.312           & {\color{red} 0.407} \\ \hline
\end{tabular}
\end{center}
}
\caption{Comparison of Results for Counter Attacks. For adversarial training, results of plain BIM attack have been reported for comparison of performance on worst suspected possible attack with our method.}
\label{tab:comp_counter}
\end{table}

\vspace{-0.2cm}
\section{Conclusion} \label{sec:conclusion}
In this work, we propose the adaptive use of Denoising AutoEncoder (DAE) for purifying adversarial attacks. We utilized reconstruction error from DAE as a base to purify adversaries by reducing it. We demonstrate how our method can be adaptively used to not just reduce reconstruction error but rather match the distribution of reconstruction errors of clean samples. We later discussed how adding random transformations and noise at each step further helps in improving the adversarial accuracy. Our framework presents a practical, effective and robust adversary purification approach in comparison to existing state-of-the-art techniques on CIFAR-10, MNIST, ImageNet. 
As a possible future work, it would be interesting to explore the use of adversary detection algorithms to provide feedback for purifying adversaries adaptively using our method.


\newpage


\bibliography{./bibliography/IEEEexample}

\newpage

\end{document}